\documentclass[a4,11pt]{report}
\usepackage{bookman}
\usepackage{textcomp}
\usepackage{graphicx}
\usepackage{fancyhdr}
\usepackage{natbib}
\usepackage[nottoc]{tocbibind}

\title{
  {\Huge\bf Data Science and Ebola} \ \\ \ \\ 
  {\large Inaugural Lecture by}}
\author{\Large Aske Plaat \ \\}
\date{\ \\ on the acceptance of the position of professor of\ \\ \ \\
  {\Large Data  Science} \ \\ \ \\ at the Universiteit Leiden\ \\ \ \\  on Monday 13
  april 2015, at 16:15 {\vskip 1.5cm} \includegraphics[width=7cm]{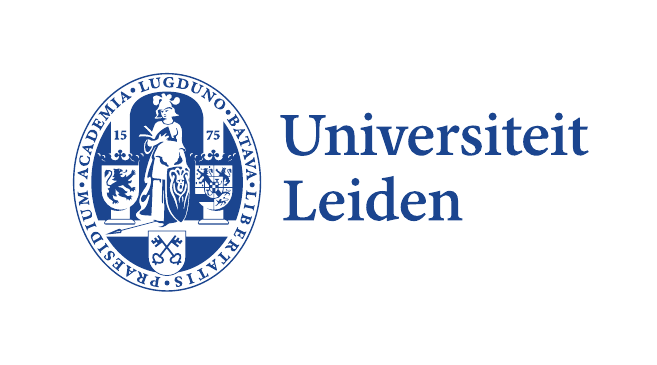}}

\begin{document}
\maketitle 

\tableofcontents
\newpage

\chapter{Addressing the Audience}
{\em Mevrouw de Rector-Magnificus,\\ 
Mijnheer de Decaan en Leden van  het bestuur van de Faculteit Wiskunde en Natuurwetenschappen,\\ 
Dames  en heren hoogleraren,\\ 
Dames en heren van de wetenschappelijke  en de ondersteunende staf,\\ 
Dames  en heren studenten,\\ 
En voorts Gij allen die deze plechtigheid met uw aanwezigheid vereert, \\

\noindent Deze aanspreektitel vormt de brug tussen het verleden en het
heden.\\
De tekst wordt in het Engels uitgesproken.}\\

\label{cha:data-science}

On 4 September
2014, shortly after the start of the academic year, the
Leiden Centre of Data Science (LCDS) was officially opened, in this
same historic building. On that day the university recognized the
importance of this new field of science. The opening speeches were by
Trevor Hastie,  professor of Mathematical Sciences at Stanford  
University, and by Prince Constantijn van Oranje, who told us about his
work for the Digital Agenda of the European Commission. Many people
have since then stressed the importance of data. By the end of 2014 both  the  European Commission and the
Netherlands government named data a key asset for the economy. Both
established a data strategy, and both announced substantial funding 
for data science research. Data may well have been the second most used word
of last year. 

Today, everybody and everything
produces data. People produce
large amounts of data in social networks and in commercial
transactions. Medical, corporate, and
government databases continue to grow. Ten years ago there were a  
billion Internet users. Now there are more than three
billion, most of whom are mobile.\footnote{source:
  http://internetlivestat.com} Sensors continue to get cheaper
and are increasingly connected, creating an {\em  Internet of
  Things}. The next 
three billion users of the Internet will not all be human, and will 
generate a large amount of data. 

In every discipline, large, diverse, and rich data sets are emerging, from
astrophysics, to the life sciences, to medicine, to the behavioral
sciences, to finance and commerce, to the humanities 
and to the arts. In every discipline people want to  
organize, analyze, optimize and understand their data to answer questions  
and to deepen insights.  

The availability of so much data and the ability to interpret it are changing the way the world
operates. The number of sciences 
using this approach is increasing. The  science that is transforming this ocean of
data into a sea of knowledge is called {\em data science}.  In many
sciences the impact on the research 
methodology is profound---some even call it a
paradigm shift.

\chapter{Ebola as Data Challenge}
First I will  address the
question of why there is so much interest in data. I will answer
this question by discussing
one of the most visible recent challenges to public
health of the moment, the  2014 Ebola outbreak in West Africa.

\section{United Nations Global Pulse}
Aid organizations recognize the necessity of
correct information for effective humanitarian aid, 
especially 
when disasters have disrupted the functioning
of  government institutions. 
The United Nations
has started Global Pulse, a flagship data science 
initiative of Secretary-General Ban Ki-moon.
The goal of Global Pulse is to accelerate discovery, development and adoption of data
science innovations for sustainable development and humanitarian
action.\footnote{http://unglobalpulse.org}  Global Pulse was started
in 2014.

The United Nations issued a report to the Secretary-General
entitled {\em A World That Counts: Mobilising the Data Revolution for
Sustainable Development}. At the presentation of the report, the
co-chair of the Expert Group, Enrico Giovannini, noted that: ``We live in a world
that faces rapidly-evolving humanitarian and development challenges,
as the Ebola epidemic so tragically proves. Governments, companies,
NGOs and individuals need good data to know where problems are, how to
fix them, and if the solutions are working. But current data are not
good enough. Too many people and issues are not counted or measured,
there are huge and growing inequalities between the information-rich
and the information-poor.''

\subsection{The 2014 Ebola Outbreak}
Outbreaks of contagious diseases lead to
severe disruptions of society, not to mention the loss of life in
all its tragedy. The history of the fight against diseases shows
some successes, such as against the plague and pox, but there are still many
diseases that we have not been able to eradicate, such as malaria,
tuberculosis, and influenza. 

Ebola is one such unsolved disease. The first reported outbreak of
Ebola was in 1976, in the rural village of Yambuku in Zaire, 100 km from
the {\em white water river}, or {\em Ebola}, as it is called in the local
tongue. The disease was so terrible, that, to avoid stigma to the
villagers of Yambuku, it was named 
after the far away river instead~(cf., \cite{wordsworth}).
Small outbreaks of Ebola have been occurring regularly in the past. They have been reported 
in Zaire and Sudan, in 1995, in 2000,  in 2003, in 2007, and in 
2012. The most recent outbreak is the 2014 outbreak, in
West Africa. This was the first outbreak in an urban environment. The outbreak started in Guinea, and spread to Liberia and
Sierra Leone. 

Researchers traced the outbreak to a two-year old child who   
died in December 2013.   
In this outbreak, half of the people who suffered from the disease died. 
As this outbreak occurred in an  
urban environment, it spread much quicker than previous outbreaks, and
caused more fatalities.   By 
early 2015 this number had reached 10,000. 
As the outbreak progressed, many hospitals, short on staff and
supplies, became overwhelmed and closed, leading health experts
to state that this may be
causing a death toll that is likely to exceed that of
the disease itself. 
Hospital workers are
especially vulnerable to catching the disease since they can easily
come into contact with
highly contagious body fluids. The World Health Organization (WHO)
reported that ten percent of the dead have been healthcare workers. 

The virus is thought to reside in fruit bats. As of this writing,
there are no approved vaccines or adequate treatments
for Ebola, although trials are under way. The disease spreads between humans by contact with
bodily fluids, such as blood, or sweat. The incubation period is long,
between one and
three weeks. 
This long incubation
period is one of the factors that allow the disease to spread so
effectively. Furthermore, Ebola
symptoms initially resemble the flu or malaria. The outbreak happened in
countries with a poor health infrastructure~\citep{Vandewalle}. Infected people are often
misdiagnosed, are not treated, and thus unknowingly infect healthy
people.

Past outbreaks were brought under control within
a few weeks; the 2014 Ebola outbreak is the first one to reach
epidemic proportions. 
The epidemic has a significant economic effect. People are   
fleeing from affected areas, creating a refugee problem and weakening   
the economy.  
Movement of people away from affected areas has disturbed agricultural  
activities.  The UN Food and Agriculture Organisation (FAO)  warned  
that the outbreak endangered harvest and food security in West  
Africa.  
Liberia and Sierra Leone struggled and initially failed to  
contain the disease. 
On 8 August 2014, the World Health  
Organization declared the outbreak  
an international emergency.


The lack of reliable data is a serious contributing factor to the 2014 
Ebola outbreak, according to the World Health Organization. 
Humanitarian aid agencies cannot respond 
appropriately; misinformation leads to widespread fear among the 
population.

\subsection{Three Data Challenges}
To address the lack of data, innovative data analysis methods can be a help. They can improve the
reliability of data, and support reducing the effects of tragedies, as the  
United Nations report on the Global Pulse indicates. In this way, data science is
changing the way that humanitarian problems are solved in our world.

As this lecture is being prepared in early 2015, aid workers and scientists have 
worked hard to contain the effects of the Ebola outbreak.
There are  three main challenges for data
scientists who are attempting to 
resolve the tragedy. These are: (1) diagnosis, (2) epidemiological
spread, and (3)
treatment and drug discovery. I will now discuss these challenges.


\section{Diagnosis}
The first challenge is in  diagnosing Ebola. Conventional diagnostic tests 
require specialised equipment and highly
trained personnel. There are few suitable testing centers in
West Africa, which leads to delays in diagnoses. In December 2014, a WHO
conference in Geneva aimed to work out which diagnostic tools could
be used to identify Ebola reliably and more quickly. The meeting  sought to
identify tests that can be used by untrained 
staff, do not require electricity or can run on batteries or solar
power and use reagents that can withstand temperatures of $40^o$C.  
On December 29, 2014, the US Food and Drug Administration approved a test
on patients with symptoms of Ebola.

\subsection{Reliable Health Data}
The difficulty in diagnosing Ebola is one of the reasons for the
disease to spread unnoticed. Doctors and hospitals  
are underequipped and therefore 
underreport Ebola cases. Months passed between the
first Ebola case and its reporting. 
Data scientists have worked to address the unreliability of Ebola data.
The Northeastern University has published an online
model which assesses the progression of the epidemic 
based on simulations of a typical epidemic spread. The analysis is
presented as a live paper that is continuously updated with new data,
projections and analysis~\citep{gomes2014assessing}. 

To acquire more reliable data, efforts have moved to crowdsourcing
initiatives that use mobile phones and SMS service. Since the SMS and voice data are
location-specific, it is possible to create maps that
correlate public sentiment to location. Others have 
created cheap  alternative diagnostic tools, such 
as checklist apps for smartphones. The 
apps may reduce fear and uncertainty among the 
population, possibly reducing the refugee problem and its disruptive effect 
on the fragile 
economies~(cf.,~\cite{gomez}).\footnote{http://www.appsagainstebola.org/} 

\subsection{Open Data, Open Government}
Knowledge is power, and governments and  organizations
are often protective of their data~(see, e.g.,
\cite{Vandewalle}). However, as the scale of the outbreak became
clear,  
governments and 
organizations started to cooperate in exchanging data on the disease. Many
organizations eventually joined open data initiatives 
that allowed scientists access to their data, to be combined with
other open data sources. 

In the midst of the fast-moving crisis, traditional
methods for solving problems did not move fast enough. Volunteer
efforts have sprung up in Africa and around the 
world in a combination of open data,
analytics software, and crowdsourcing. IBM has set up an African Open Data
Initiative to help African 
countries tap open data as a means of addressing health,
infrastructure and economic challenges. The World Health
Organization provided data. In New York a
grassroots Ebola Open Data Jam was
organized.\footnote{http://eboladata.org} The UN Office for the
Coordination of Humanitarian Affairs set up a Humanitarian Data
Exchange.\footnote{https://data.hdx.rwlabs.org} The government of
Sierra Leone created its own Open Data
initiative.\footnote{http://www.ogi.gov.sl} The Ebola epidemic caused
the Liberia Government to provide data on their government to the
outside. In this way it faciltated the step towards Open
Government.\footnote{http://www.opengovpartnership.org/country/liberia} 

These open data initiatives are of great value since they allow
different scientists to work on the data, to combine data sources, and
to  improve their models. For example, one of the 
findings of the project for the Global Data on Events, Location and Tone 
(GDELT) is that a global monitoring of internet and media news can
provide a 
picture of the unfolding of the outbreak that is as accurate as
ground truth
data,\footnote{https://keystoneaccountability.wordpress.com/tag/nick-van-praag/}
only much faster.\footnote{http://gdeltproject.org,} 

The availability of different data sources allows data
to be triangulated, or cross-checked, which improves data quality. 
Models have been  made to visualize the spread of the disease using
heat maps that correlate locations to public sentiment, migration,
infections, and fatalities.
Special tools, such as the Spatiotemporal Epidemiological Modeler tool,
are designed to 
help scientists and public health officials create real time models of emerging
infectious diseases.\footnote{https://www.eclipse.org/stem/} 

\section{Epidemiological Spread}
\label{sec:migration}
The second challenge is how to model reliably the spread of the
epidemic. Epidemiologists traditionally have to rely on anecdotal information,
on-the-ground 
surveys, and police and hospital reports. This type of data is often collected too
slowly to curb the spread of the disease. Scientists have been working
under time pressure to develop
novel methods to map the spread more quickly and more precisely. Below
I discuss two methods, viz.\ analyzing mobile phone data
(subsection~\ref{sec:mobile}) 
and improving contact tracing  (subsection~\ref{sec:tracing}).

\subsection{Mobile Phone Data}
\label{sec:mobile}\label{sec:millions}
The first method is to analyze mobile phone data. Mobile phones are
nowadays widely owned in even the poorest countries in 
Africa. They are a rich source of data in a region where only a few other
reliable sources are available. 
Orange Telecom in Senegal handed over anonymized voice and text data
from 150,000 mobile phones to a Swedish non-profit organization, whose
data analysts drew up detailed maps of typical
population movements in the region. 
Authorities and aid workers could then see where the best places were
to set up treatment centers. Authorities also used this information
to find the most effective ways to restrict travel in an
attempt to contain the disease. 

A second way in which phone data is used, is by
tracking the number of calls to helplines. 
A sharp increase from one particular area could
suggest an outbreak and alert authorities to direct more resources
to that area. Software companies are helping to visualize this data and
overlay other existing sources of data from ground truth data to build up a
richer picture. 

Mobile phone data can be used to improve the accuracy of
epidemiological models. Epidemiology
uses advanced statistics to 
model the spread of a disease, often based on historical data, the level
of contagiousness of
the disease, and on behavioral factors. 
Dynamic models combine historical data with current field
measurements. The dynamic models can be more precise in their
prediction of the spread of a disease than static historical
models. The difference in accuracy can 
be large, with serious consequences for policy makers. 

As a case in point, we mention a
report of 
September 2014 by the Center of Disease Control. It analyzes the impact of underreporting
and suggests correction of case numbers by a 
factor of up to 2.5. With 
this correction factor, approximately 21,000 total cases were
estimated for the end of September 2014 in Liberia and Sierra Leone
alone. The same report predicts that the number of cases could reach
1.4 million in Liberia and Sierra Leone 
by the end of January. Two months later, at a congressional
hearing, the director of the CDC said that the number
of Ebola cases was no longer expected to exceed 1 million, moving away
from the worst-case scenario that had been previously predicted. 

New data allow new mathematical models to be validated. One model that 
has attracted attention is the IDEA model, a straightforward two parameter 
mathematical model that appears to model the spread of the disease
well~(cf.,~\cite{fisman2014early}). 
 
Access to real time data, such as the measurement of migration
patterns through mobile phone tracking, is of great value to
improve epidemiological 
models. Incorporating data from different sources into simulation
models allows data triangulation 
to predict the spread of the disease better.  
Improving 
the accuracy of statistical models is important not only for better 
targeting of relief work, but also for improving the reputation of aid 
organizations as providers of trustworthy information.

\subsection{Contact Tracing}
\label{sec:tracing}
The second method for better mapping the spread of the disease is to improve
contact tracing. Contact tracing is an 
important  method (1) for understanding the spread 
of Ebola and (2) for acquiring correct numbers on the size of the 
epidemic. Contact tracing requires effective community surveillance so
that a  possible case of Ebola can be registered and
diagnosed. Subsequently everyone who has had close   
contact with the patient must be found and tracked for 21 days. This 
requires careful record keeping and many properly trained and equipped 
staff. 
There is a substantial effort to 
train volunteers and health workers, sponsored by USAID. According to 
WHO reports, 25,926 contacts from Guinea, 35,183 from Liberia and 
104,454 from Sierra Leone were listed and being traced at the end of
2014. 

Contact tracing is labor intensive. Patients are
interviewed and their relatives over the past period are contacted to
establish how they were likely infected, and whom they could have
likely infected. Contacts are watched for 21 days, to see whether they
develop symptoms of the illness. Thus, a social graph of the patient is built.
By combining social graphs of people in an area an overall view of the 
network of the disease in a certain area and  time period can be
created.\footnote{CDC Methods for Implementing and Managing Contact
    Tracing for Ebola Virus Disease in Less-Affected Countries,
    Centers for Control \& Prevention, 2014} 

Estimating the spread of the disease is difficult.
A study published in December 2014 by~\cite{scarpino2014epidemiological} found that transmission of the 
Ebola virus occurs principally within families, in hospitals and at 
funerals. The data, gathered during three weeks of contact tracing  showed that the third person in any transmission chain often 
knew both the first and second person. The authors estimated that 
between 17 percent and 70 percent of the cases in West Africa are 
unreported. Prior projections had estimated a much higher figure.  The study concludes that the epidemic is not as difficult 
to control as feared, if rapid, vigorous contact tracing and 
quarantines are employed. 

Traditional contact tracing methods involve traveling to patients and
interviewing them. Online social networks and contact lists of
patients  provide quick information about the kind of network and
travel patterns of patients. Patients with many contacts 
and an active travel pattern can be quickly identified,
allowing more efficient use of scarce tracing personnel. Current manuals do not prescribe taking online
information into account. Apps are
being developed to ease the process of contact tracing.\footnote{http://www.appsagainstebola.org}
We conclude that innovative contact tracing methods such as analyzing
online social networks, mobile phone data and apps can speedup the process of
contact tracing, to better map the epidemiological spread.

\section{Treatment \& Drug Discovery}
The  third challenge that I will discuss is related
to the prevention of
Ebola. It concerns treatment and drug discovery. Pharmacologists have
developed a range of high 
performance drug discovery techniques over the past years. They are used
intensively to find a cure for Ebola.

\subsection{Treatment}
At the time these words were written there is no approved vaccine for
Ebola, despite a large effort by the pharmaceutical industry. In
addition, there is no
cure or specific treatment that is currently approved. Treatment is
primarily supportive in nature, as
survival chances are improved by early care with rehydration and
symptomatic treatment. A
number of experimental treatments are being considered for use in the
context of this outbreak, and are currently in
clinical trials. 
Patient data is recorded to understand the most effective combination
of therapies. In other diseases a well balanced combination of
symptomatic treatment has been shown to increase both  life expectancy
and the quality of life of patients. 
Transparent access to reliable patient records for doctors and
scientists is necessary for effective treatment development.

\subsection{Drug Discovery}
Finding a preventive vaccine for Ebola is of prime importance.
According to a recent study by the US National Institute of Allergy and Infectious
Diseases the scientific community is still in the
early stages of understanding how infection with the Ebola virus can
be treated and prevented.
Many Ebola vaccine candidates have been developed in the decade prior
to 2014, but none has yet been approved for clinical use in
humans. Several promising vaccine candidates protect nonhuman primates (usually macaques) against lethal
infection, and some are now going through the clinical trial process.

The process of drug discovery has advanced to a state
where many steps 
have been automated. High throughput screening is a method for
scientific experimentation in drug discovery. It uses robotics, data
processing and control software, and includes sensitive detectors and
devices for handling liquids. High throughput screening allows a
researcher to 
conduct quickly millions of chemical, genetic, or pharmacological
tests.  Researchers have  developed
computational methods to analyze these test results.

Results from high throughput screening are used to refine
simulation models of the virus, in order to design a
vaccine. Simulation data can then be checked with in-vitro
observations.  

Pharmacology and molecular biology are  active fields of research,
where many results on gene-disease findings and related
drugs are published. In addition to analyzing {\em databases\/} of molecules
and proteines the {\em publications\/} themselves 
allow a drug discovery method based on text mining and
statistics. In this method textual
correlations in scientific papers 
are analyzed. A high textual correlation indicates an increased possibility of a
relation between molecules and diseases, warranting further
research. The advantage of such in-silico drug discovery are (1) the low
cost and (2) the systematic nature of the search,
allowing a much wider investigation of acceptable relations than is possible
with traditional methods.

\section{Ebola: Three Challenges for Data Science}
At this point, we have discussed three challenges where data science is helping
to resolve the Ebola tragedy. The outbreak occurred in countries with
a poor health infrastructure, and a lack of reliable
data. Governments and organizations learned the importance of opening up
 their data. Data scientists could then work on (1) better methods for diagnosis, (2)
new online epidemiological models, and (3) developing 
vaccines and treatment methods. 

Ad (1) Open data initiatives improve the quality of data about the
outbreak. 
%
Novel 
methods such as smartphone self assessment apps have been developed, and
the movement of people is analyzed 
based on data from mobile phones. 

Ad (2) New online epidemiological
models are developed that help simulate 
the spread of the disease based on data that is continuously being
updated.
A relatively new area is the analysis of online social networks and
call information for contact tracing, to improve the accuracy and efficiency of manual
methods.


Ad (3) Pharmacologists are working hard to develop vaccines and
treatment drugs for Ebola, making use of high
throughput drug discovery methods and data analysis in trials.\\

\noindent In conclusion, data science has permeated the methods of doctors, aid workers,
epidemiologists, and pharmacologists, helping them to fight the
disease. 

Let us now look into more detail at the
technologies that data science is using. It allows us to understand
future developments for Ebola, and for other domains.



\chapter{Data Science Technologies}
%
In the past, data collection and processing  techniques were limited in
their power and versatility. In the last decade techniques have
progressed considerably. For Ebola a wide range
of data sources are used, such as mobile phone data, diagnostic app data,
social network data,
and advanced mathematical models. Combining these kinds of data requires
new data processing technologies. 

We will now describe three techniques in
more detail.
(1) For {\em diagnosis\/} we will look at data
quality and representation techniques, (2) for {\em epidemiological spread\/} we will look at
analysis techniques for diverse and large data sets, and (3) for {\em treatment\/} we will look at
high performance data analysis techniques.

\section{Data Quality \& Representation}
I will start with techniques for data quality and representation that are
used in the diagnosis part of the Ebola outbreak.  

\subsection{Quality of Ebola Data}
An important aspect of data science is data quality. In many projects
the most time consuming 
task is ensuring the quality of the data: cleaning the data, and checking
for missing and inconsistent values~(see, e.g., \cite{rahm2000data}). For Ebola,
gathering high quality data is  a difficult challenge, and
alternative sources were sought, such as mobile phone data and
internet news postings. These additional sources allow data to be
triangulated so that the quality increases. Also, data can be
collected more quickly and is broader in
scope.


\subsection{Knowledge Representation Techniques for Ebola}
In diagnosing Ebola data from different sources is collected in
different data sets. It is in combining 
data from different areas where the real power
of data science lies: triangulating data to improve data quality, and also, finding
unexpected patterns. To be able to 
compare items from different data sets, the data must be represented
in an organized and comparable manner. The field of knowledge
representation studies 
this aspect. It uses techniques such as semantic networks and automated
inferencing to organize knowledge in taxonomies and ontologies. Semantic web
techniques for linked open data allow automatic inference of diverse
kinds of data, such as social network data~(cf.,~\cite{groth2012semantic}). 
Social and semantic network techniques
are areas of active research. Their use in helping to diagnose Ebola
cases illustrates how fundamental research and real world challenges
can go together. 
 
\section{Analysis Techniques for Diverse \& Large Data Sets}
I will now discuss two techniques for analysis of diverse and
large data sets that are 
used in modeling the epidemiological spread of the Ebola outbreak. 

\subsection{Diverse Data Sets}
Epidemiology makes good use of statistics and data analysis techniques. Developers  of
statistical methods have a  
 history of standardizing their best algorithms into libraries and 
 software packages. 
Well known packages that are used in epidemiology 
 are SPSS~\citep{meulman2001spss,field2009discovering}, 
 Weka~\citep{witten11:_data_minin}, and R~\citep{team2012r}.   These packages
have paved the way for the use of data
analysis techniques in epidemiology and in other sciences. 



The data sources that are used for tracking the spread of the 2014
Ebola outbreak are 
diverse and go beyond traditional tables of numerical data. Data can
be text documents, sound, pictures, even
video, and data from motion sensors. Data can be dynamic, for example an
incoming stream of messages or video. 
Conventional, linear, statistical methods are not suited to analyze the 
data from the Ebola outbreak. Efforts to analyze this kind of high
dimensional data have yielded new statistical and machine learning techniques
(see, e.g., \cite{hastie2009elements,johnstone,takes2014algorithms,schraagen}). 
As the Ebola case shows, still more techniques are needed, and
the advanced techniques must be packaged in a way that is accessible
for  epidemiologists and other scientists. 

\subsection{Large Data Sets}
Current data sets are larger than before, have a more diverse
structure than before, and  change more  
frequently than before. Finding answers in such large, unstructured, data sets
requires intelligent search algorithms that adapt to the search space
at hand. Many years ago, in Rotterdam and Edmonton, I started to work
in this field, as part of my PhD research.

For analyzing large  data sets a variety of
adaptive search
techniques exists, ranging from 
stochastic methods~(see, e.g., \cite{hoos2004stochastic,ruijl2014local,ruijl2014hepgame}), multiple-objective
optimization~(see, e.g., \cite{koch2015efficient}), evolutionary
algorithms~(see, e.g.,
\cite{back2013contemporary,back2014introduction}), to new versions of
neural networks~(see, e.g., \cite{krizhevsky2012imagenet}). These
techniques have shown 
remarkable success, although many challenges remain. 
In
subsection~\ref{sec:future} I will describe ideas for future research.

\section{High Performance Drug Discovery Techniques}
In searching for vaccines, high performance data analysis techniques are
heavily used. I will briefly discuss techniques from high performance
computing, a field in which I worked as a postdoc, first in Cambridge at MIT, and
later in Amsterdam at the VU.

Quickly analyzing large data sets requires fast algorithms and fast
computers. 
Initially supercomputers were used for numerical modeling,
for applications such as computational fluid dynamics, for weather
prediction, and for simulations ranging from  nuclear to galactic
processes. 

In contrast, many of the drug discovery techniques for Ebola involve
classification and discrete choice (both
for epidemiology and for vaccine discovery). These problems require
the application of
combinatorial methods, as used, for
example, in route 
planning  problems, scheduling (see, e.g., \cite{plaat1996research,hoos2004stochastic}), or for searching for relations between genes and
diseases in large databases.  
Together, methods from  
numerical and combinatorial analysis comprise data science.
There have been great advances in high performance computing, combinatorial
optimization, and databases (see, e.g.,
\cite{plaat2001sensitivity,boncz2008breaking,dean2008mapreduce,seinstra2011jungle,Engle:2012,mirsoleimani2014performance}).
These have enabled the  
application of supercomputing to fields as diverse as the life
sciences, the social sciences, and the humanities.

Due to the increased need for data analysis the worldwide demand for compute
power is increasing sharply. In this respect it is remarkable to see that the
Netherlands investment in scientific compute power is not keeping
pace. Our place in the worldwide list of
supercomputers, the TOP 500, is embarrassingly low, and certainly not
commensurate with that of a data economy.\footnote{http://top500.org} 




\section{Data Science: Three Techniques for Ebola}
We have discussed three techniques that are used to resolve the 
Ebola tragedy. These are techniques for (1) collecting high quality data and 
organizing the data so that combinations between 
data sets of a diverse structure can be made, (2) for the 
analysis of large and diverse data sets, using adaptive techniques 
for high dimensional data sets, and (3)  high
performance drug discovery techniques. High performance techniques are 
necessary since the size of the 
data, especially when combinations are made, quickly becomes too large
for ordinary computers.


\chapter{Multidisciplinary Cooperation}
We have now surveyed data science  techniques that
are used for Ebola and that have changed the way in which the disease
is handled. For a moment we will digress
and  look at other applications, outside the life sciences, in which data science is
causing a similar change. We will start with astronomy.

\section{Astronomy}
In  astronomy, the Low Frequency Array (LOFAR) radio telescope consists of 25,000 small
antennas that are spread out over a larger area to effectively form
one large virtual antenna~\citep{rottgering2006lofar,van2013lofar}. LOFAR's antennas
together  generate so much raw data that it has to 
be reduced before it can be stored for further  
processing and analysis~(cf., \cite{de2009lofar}).  A dedicated
supercomputer, BlueGene/L, has been built to do the 
signal processing of LOFAR~\citep{Romein:06,Romein:11a}. LOFAR's
design has not only been made possible by advanced sensor 
technology, but also by fast signal processing algorithms and
large compute power.


\section{Physics}
In physics, particle experiments generate large amounts of
data. On the 4th of July of 2012 one of the most important scientific discoveries in physics
was announced: two independent experiments reported results that were
consistent with the detection of the Higgs 
boson~\citep{aad2012observation,chatrchyan2012observation}, the last elusive particle from the standard model. A year 
later Peter Higgs was
awarded the Nobel prize, together with Francois
Englert.\footnote{http://press.web.cern.ch/press-releases/2011/12/atlas-and-cms-experiments-present-higgs-search-status}
Calculations from almost 50 years ago, predicting the particle's
existence, had been proven correct. 

It has been reported that around 100
Petabyte of data has been generated in the Large Hadron Collider at
CERN in these experiments. 
To put that amount in perspective, my somewhat older laptop has a storage capacity of
128 Gigabyte. The amount of data stored at CERN would
require 800,000 of those laptops to store it. 

In addition to experimentalists, theorists too work with large
amounts of data. Ever since the 1960s theoretical physicists have
been using computers to manipulate large formulas to predict
experimental results. Veltman and 't Hooft used a special computer algebra system for the
calculation for which they received their 
Nobel prize in 1999. Inspired by their system Jos Vermaseren 
developed an improved system called FORM,  to work with such large
formulas~\citep{vermaseren2000new,ueda2014recent}.\footnote{http://www.nikhef.nl/\~{}form}

For the next generation of particle experiments even more complex
  calculations are 
required. In the HEPGAME project, for which we gratefully
acknowledge EU funding through the ERC Advanced 
scheme, Jos Vermaseren, Jaap van den
Herik and I work, with our PhD students and postdocs, on advanced combinatorics and physics
to make these complex calculations in FORM
possible~\citep{vermaseren,ruijl2014hepgame,mirsoleimani2014performance}.\footnote{http://hepgame.org}



\section{Law Enforcement}
One field with a natural interest in the behavior of their
``customers'' is law enforcement. Activities to reduce 
terrorism, crime, hooliganism, and jihadism are becoming increasingly driven by
information. 

Data-driven
methods are credited with 
modern policing successes in Los Angeles, New York, and other
cities.\footnote{http://www.governing.com/topics/public-justice-safety/Data-driven-Policing.html} Our national police is also 
gathering data to increase effectiveness, by correlating crime figures with police actions.
{\em Intelligent\/} blue, instead of 
more blue, is the new motto.
Many police forces are experimenting with intelligence led
policing~\citep{meesters}. Other data-driven methods have also shown 
success. Combining scenario methods with data analytics can be
used to anticipate criminal behavior to some degree~\Citep{kock}.

\section{Commerce}
Data science is an important factor in the 
online and offline economy.
In bookselling (Amazon) and online video (YouTube, Netflix) the volume of buying decisions and
views allows
statistically significant personalized recommendations to be computed.
These recommendations  drive
much of  sales. 
Data warehouses have become core systems, for example, 
for calculating online ticket prices in the hospitality and travel
industry. 

\section{Regulation}
Increasingly we live our lives online, where  
expectations about privacy may not hold. 
Technology is proving 
a difficult topic for    
regulators that wish to protect our rights. As in all of our society, moral and ethical issues arise, and 
research into the philosophical and legal aspects of behavioral data 
collection is an important area (see, e.g.,
~\cite{bibivandenberg,FM4010,van2014plagiarism}).   Active legal 
research is needed, and legal scholars   
need to have an adequate technological  
understanding~\citep{prins,vdbergkeymolen,van2014technology}.

\chapter{Outlook}
Having looked at how data science is used for Ebola and other
applications, it is now time to look into the 
future. First we will discuss how data science improves the chances
of preventing future virus outbreaks. Next we will discuss plans
for data science
research. Finally, we will look at data science in Leiden.

\section{Future Ebola Outbreaks}
The medical history of conquering diseases is one of many successes,
although important challenges remain.

The loss of some 10,000 lives since the 2014 Ebola outbreak and the ensuing
human and social disruption are deeply tragic. So tragic, and so large
is the impact of the outbreak, that it
caused scientists, volunteer efforts, and the international community
(1) to work around the clock to improve diagnosis techniques for the
disease, (2) to improve
the tracking of the 
spread of the disease, (3) to gather information on patient treatment,
and (4) to work on experimental vaccines. 

Due to the urgency of the situation
governments and organizations opened up their data, and
researchers used the latest data science 
techniques in their approaches. Regardless of the reason novel
medical and data science methods 
were created. 
The accuracy of epidemiological models improved
greatly---causing predictions to be adjusted by more than an order of
magnitude  (see section \ref{sec:millions}).  
Pharmacologists have been  
working hard to create vaccines. At the time of this writing the first
vaccines have been scheduled for phase 3 trials at the end of 2015.

Unless vaccines will become available at a low cost and at a wide scale, it
is likely that Ebola incidents will continue to occur. However, with
open governments and
the concerted effort from the international scientific community, new
diagnostic tools, epidemiological models, and treatment and drug discovery
methods
have been developed. Together these will likely prevent outbreaks of
the scale of 2014. 
Many of the lessons, such as advances in dynamic epidemiological
modeling, are likely to help in containing other diseases as well. 

The availability of open data and data science technology are causing
fundamental changes 
in our science and in our society, as the Ebola case illustrates. Data
science methods have become indispensable in containing  
such an outbreak.  
The simple fact that significantly more data can be measured,
analyzed, and understood, has profound 
implications in all of science and society.



\section{Future Developments in Data Science}
The data science vision is  to measure more, to analyze 
more, and to know more. It is used in policy making:
evidence-based-policy-making should lead to better  
decisions. It is used in policing: Intelligence-lead policing will reduce crime. Better data
on consumer preferences allows marketers to create better
recommendations and advertisements. More data allows astronomers to
uncover more about  our universe. By using large scale text analysis
historians can better understand historical developments.  The
Internet of Things will cause a revolution in predictive maintenance.
When the right information is available humanitarian aid can be more
effective. New drugs can be discovered, outbreaks of diseases 
can be stopped, and diseases may eventually be eradicated, when the
right information can be gleaned from the data.  

Realizing this vision is dependent on science and  technology. We must
be able: (1) to have high quality data that can be represented and
combined in a meaningful manner, (2) to analyze diverse and large data sets, and (3)
to do so quickly, using high performance computing methods.

The goal of my chair is to
create new data science methods, 
for the large and diverse data sets that scientists
are increasingly using. 
%
%
Let us now look in more detail at the improvements that we will be
working on. We will start with
data quality and representation.

\subsection{Data Quality \& Representation}
%
Open scientific data sources enable new forms of
knowledge discovery. Publication mining experiments
can yield results 
at significantly lower cost than traditional {\em in-vitro\/}
experiments. The work in the field of 
knowledge representation, which studies taxonomies and information
classification, is of great importance here. At our center we are
working on  extending this
to other sciences, for example to text
mining of historical texts and to crowd sourcing for
museum collections. Promising cooperations between 
biosemantics groups, database groups and high performance
computing groups are happening. 


We are in the fortunate position to have as
part of our center one
of the leading high performance database systems researchers, who
works on
MonetDB~\citep{boncz2008breaking}. This is a
great asset for all our data management research projects.

\subsection{Analysis Techniques for Diverse \& Large Data Sets}
Advances in modeling drive statistical and computational techniques
for Ebola, and for data science. An example is the need for validation of simulation
results of multi-scale models~(see, e.g., \cite{portegies2013multi,merks2015}).  
The increased complexity of these models will demand better
validation methods and lead to an
increased need for observation data for initial values. In our center
we are working on machine analysis of numerical simulation data, and
on predictive maintenance. 
To compare the quality of algorithms benchmarks are needed. In machine
learning we see benchmarking initiatives for data sets and algorithms, such as UCI\footnote{https://archive.ics.uci.edu/ml/datasets.html} and
OpenML.\footnote{http://www.openml.org} We will use OpenML in our
 algorithm development work.

\label{sec:future}
In many research projects a wide range of real world data is used from health (such as
Ebola), to vibration data from bridges, to financial data
from mortgages and commerce. LUMC, IBL and LIACS are working on knowledge
representation and combinatorial 
optimization for metagenomics applications. 
Combinatorial
algorithms are fruitfully applied in logistics
operations in humanitarian aid, and many other
scientific applications that have planning and scheduling challenges.
In high performance combinatorial optimization a fundamental move
to parallel algorithms has occurred, for example in high throughput
drug discovery. This creates challenges for
algorithm designers and a strong need for formal verification
methods. I am looking
forward to cooperation in this area. 

At a more fundamental level, there is exciting work happening in search space  
analysis and visualization~(see, e.g.,
\cite{verbeek2007visualizing,ochoa2014local}), a possible combination
with solution trees is
interesting~\citep{plaat1994nearly,plaat2014sss}.   
We will look for relations between natural and heuristic optimization:
finding common elements in evolutionary
approaches~\citep{back2014introduction}, deep neural
nets~\citep{krizhevsky2012imagenet,van1996neural}, pattern
recognition, stochastic search~\citep{kocsis,kuipers2013improving,ruijl2014local}, and
classical  enumeration algorithms~(see, e.g.,
\cite{russellartificial,ruijl2014local,ruijl2014hepgame}). Recent experience with
deep neural nets and stochastic optimization suggest the feasibility
of such relations~(see, e.g., \cite{maddison2014move,clark2014teaching,Mnih2015}). Also, genetic and
evolutionary algorithms are often 
successfully applied to classic optimization
problems~\citep{Izzo:2013:SGT:2463372.2463524,6626616}. 

Such new methods  will allow even more complex  problems to
be analyzed  and understood.
Many of the successes are driven by real world data
and real world problems, such as the Ebola outbreak. 
Promising application areas are the analysis of
simulation output, predictive maintenance, and drug discovery.
Further application fields are humanitarian aid, marketing,
organizational behavior and management (see, e.g., \cite{plaat2010vlinder}),
financial analysis, and 
sports and, of course,  health.

\subsection{High Performance Computing}
\label{sec:hpc} 
High performance computing in Leiden has a rich history in areas such as
numerical computing, compiler technology,  
embedded systems, distributed computing, and sparse matrix
codes with strong groups throughout the Faculty of Science. The work
in the DAS projects~\citep{seinstra2011jungle}, 
and the Little Green Machine\footnote{http://www.littlegreenmachine.org} is state
of the art. Data science needs a strong high performance systems group and we will work
to further strengthen this field. Among the research topics are
questions in compiler technology, data 
sharing (see, e.g., \cite{kielmann1999magpie,plaat2001sensitivity}),
work and data scheduling~(see, e.g., \cite{romein2002performance,kishimoto2013evaluation}), accelerators such as GPGPU
(see, e.g., \cite{mirsoleimani2014performance,karami2014statistical}),
and other topics. Many applications, for example, imaging and
astrophysics simulations, can benefit from 
these methods. 

 

\section{The Leiden Centre of Data Science}
Scientists and society have
found that high performance analysis methods
can solve some of their problems and answer some of their questions. 
In fields ranging from the humanities, to 
astronomy, to containing outbreaks of contagious diseases, they
learn more and create new insights. 

The purpose of the Leiden Centre of Data Science is to solve real
world problems in science and society. In the 
process, these efforts drive the invention of new data science
techniques. 
The interest in data science is high, inside our university, and
outside. The Leiden Centre of Data Science is an ideal network
organization for
cooperation with other universities, academies, and data 
science centers. We cooperate with national and local governments,
with commercial companies, and with social and cultural institutes.  
We organize data science summer schools, labs, and regular academic courses.

The purpose of the Centre is also to facilitate cooperation between
different disciplines. As such our focus is on community building,
on building a research infrastructure, and on initiating
projects with researchers in academia and 
industry. Since the official opening of the Centre, less than a year
ago, much has been achieved, and the interest in data science has only
grown.

\section{Conclusion}
Mathematics and computer science are impacting our modern lives in many  
ways. 
The growing availability of data and data processing technology is causing profound changes in
science and society: from the way that
the Ebola outbreak is approached, to predictive maintenance of
bridges and infrastructure, to fine grained marketing, to finding new 
drugs, to monitoring health
behavior, to finding social networks in ancient Chinese 
texts, to analyzing computational fluid dynamics
simulations, and to large scale
simulations of star systems. Patterns are everywhere, and by learning
to find them in large and diverse data sets we gain insights and
solve problems. 

From the perspective of data science the 2014 Ebola outbreak is of
special significance. Governments and organizations provided open
access to data, enabling the active creation of new data science
tools. New diagnosis techniques and  
better epidemiological models have succeeded in  
reducing an aggressive epidemic faster than would otherwise have been possible. 
Pharmacologists are using high throughput  
methods that are increasing the chances of finding a vaccine.  
As a result, there is legitimate hope that this epidemic is soon under control. 
Data science is helping to save many lives---abstract notions from the worlds of statistics and
algorithms are having an effect that is anything but abstract. The
world is full of data, and data scientists are here to help.

At the start of this lecture I asked the question why there is so much
interest in data. Subsequently, we discussed some important practical
examples. Kurt Lewin once remarked that there is nothing so practical
as a good theory, and I agree. For data science, such a theory has
been proposed six years ago. In 2009 data-intensive research methods
were named {\em the Fourth Paradigm} \citep{hey2009fourth}. This term implies
that data science complements the methods of induction, deduction, and
simulation, the three paradigms of the experimental cycle. The term  
suggests that data science is as fundamental as these  
scientific paradigms.  

Data science gives scientists a new, disruptive, way to look for
answers to their questions. For its consequences in theory and practice, we welcome data as a new  
paradigm to science.  

\chapter{Acknowledgments}
We have come to the end of this lecture. So, I would like to
express my sincere thanks to  the people who have been instrumental in the creation
of this chair. 

First of all, I thank the Board of Leiden University and the Board of
the Faculty of Science for appointing me as professor of data
science. I am grateful to the Boards, and in particular to Carel Stolker and Geert de Snoo, for
supporting and creating the Leiden Centre of Data Science. 

Joost Kok, scientific director of the Leiden Institute of Advanced
Computer Science and head of the data mining group, was instrumental
in the creation of the Centre and of this chair. Joost, I thank you for
leading this wonderful Institute and for discussing so many ideas and  insights.

Jaap van den Herik is an extraordinary man. Jaap, we first met twenty-one years
ago. Working with you is wonderful. I thank you for your support,
your energy, your honesty, your wisdom, your help, 
and for your friendship. Leiden is truly privileged to count you among its
professors. 

I thank Johanna Hellemons, longtime manager of the group I had the
good fortune to join. Johanna, you are our rock
and our tower of strength.

I would like to thank all the wonderful
people that I have worked with and that make science such a great
adventure every day. Let me mention six people. I thank Jos
Vermaseren for his visionary idea to join the 
worlds of physics and artificial intelligence. It worked out
wonderfully.  At LIACS I met Thomas B{\"a}ck, Bernard Katzy, Walter Kosters, Hans
le Fever, and 
Katy Wolstencroft. Thank you, and all
the people that I cannot mention individually, for everything!

For their help in Ebola research I thank Robert Kirkpatrick from UN Global Pulse, I thank Abdul
Hafiz Koroma from the Ministry of Public Works of Liberia, I thank Uli Mans,
Gideon Shimshon and colleagues from the Leiden Centre for Innovation,
I thank Thomas Helling, Mirjam van Reisen and Bartel van de Walle, I thank Meenal Pore from
IBM Africa, and I thank Philips Research. 

I thank my students, who have taught me so much. Science is a people's
business, which is what I like about it so much (in addition to
finding out new things, which is also very cool). 

The last few words of this lecture are for three people, whom I value
even above my love of science: Rosalin, Isabel, and Saskia. I thank
you, with all my heart, for your love and laugther
through all these years. \\

\noindent {\em IK HEB GEZEGD.}

\bibliographystyle{apalike2}

\renewcommand\bibname{References}
{\small \bibliography{biboratie}}

\end{document}